%% file: main_11_11.tex
\documentclass[10pt,twocolumn,letterpaper]{article}

\usepackage[pagenumbers]{cvpr} 

\input{preamble}
\definecolor{cvprblue}{rgb}{0.21,0.49,0.74}
\usepackage[pagebackref,breaklinks,colorlinks,allcolors=cvprblue]{hyperref}
\usepackage{booktabs,adjustbox}
\usepackage[table]{xcolor}
\usepackage{pifont} 
\usepackage{multirow} 
\usepackage{dsfont}
\usepackage{algorithm}
\usepackage{algpseudocode}
\usepackage{amsmath}
\usepackage{amssymb}

\def\paperID{***} 
\def\confName{CVPR}
\def\confYear{2027}

\title{Remember, Replay, Adapt: Learning from Past Experience \\ in Online Vision-and-Language Navigation}

\author{Chenhao Ding\\
The University of Tokyo\\
Tokyo, Japan\\
\and
Jizhou Han\\
Xi'an Jiaotong University\\
Xi'an, China\\
\and Lixin Xiu\\
The University of Tokyo\\
Tokyo, Japan\\
\and Muyao Wang\\
The University of Tokyo\\
Tokyo,Japan\\
\and Songlin Dong\\
Shenzhen University of \\ Advanced Technology\\
Shenzhen, China\\
\and Hideki Nakayama\\
The University of Tokyo\\
Tokyo, Japan\\
}

\begin{document}
\maketitle

\input{sec/0_abstract_11_11}

\input{sec/1_intro_11_11}
\input{sec/2_related_work_11_11}
\input{sec/3_method_11_11}
\input{sec/4_experiment_11_11}
\input{sec/5_conclusion_11_11}
{
    \small
    \bibliographystyle{ieeenat_fullname}
    \bibliography{main}
}


\end{document}

%% file: sec/0_abstract_11_11.tex
\begin{abstract}
Existing feedback-based methods for online vision-and-language navigation (VLN) primarily learn from current episode outcomes, leaving previously acquired feedback underutilized. Inspired by human episodic memory, we propose \textbf{R}$^2$\textbf{A}, an online VLN adaptation framework that continually reuses successful and failed navigation experiences for policy learning. Specifically, \textbf{S}uccess–\textbf{F}ailure \textbf{M}emory~(\textbf{SFM}) retains successful and failed navigation episodes with acquired feedback and retrieves task-relevant experiences based on incoming instructions, enabling historical feedback to guide subsequent navigation. \textbf{P}rior-\textbf{A}nchored \textbf{A}daptation (\textbf{PAA}) learns from these experiences by reinforcing successful trajectories and suppressing failed ones relative to a frozen pretrained policy, while regularizing behavioral drift.
Together, these modules enable continual feedback reuse and stable policy adaptation without source-domain data or expert supervision. Extensive experiments across three benchmarks, REVERIE, R2R, and R2R-CE, and five navigation backbones demonstrate consistent performance gains. On the REVERIE validation unseen split, \textbf{R}$^2$\textbf{A} improves DUET's SR from 46.98\% to \textbf{69.84\%} and SPL from 33.73\% to \textbf{50.98\%}.
\end{abstract}

%% file: sec/1_intro_11_11.tex
\begin{figure*}[t]
    \centering
    \includegraphics[width=\linewidth]{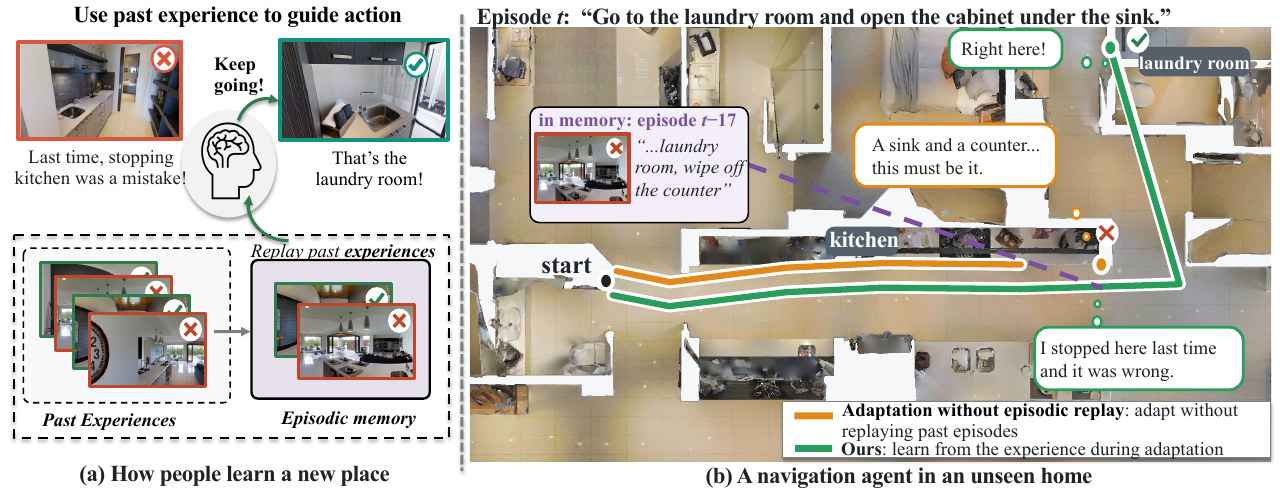}
\caption{\textbf{Learning from past successes and failures for navigation.} (a) People recall successful experiences and mistakes to guide navigation. (b) Similar visual cues may cause an agent to confuse a kitchen with a laundry room; replaying a past failure helps correct this mistake. Inspired by episodic memory, R$^2$A reuses successful and failed experiences with outcome feedback for adaptation.}
\label{fig:motivation}
\vspace{-2mm}
    \label{fig:head}
\end{figure*}

\section{Introduction}
\label{sec:1intro}

Vision-and-language navigation (VLN)~\cite{vln} requires embodied agents to follow natural language instructions to navigate visual environments. During deployment, unfamiliar spatial layouts, visual appearances, and object–room associations challenge navigation policies trained on offline data~\cite{lanvln,streamvln,navgpt,navgpt2,frevln}. Test-time adaptation~\cite{tent,tpt} enables agents to improve navigation by learning from experience accumulated during deployment. As agents undertake successive tasks, their trajectories and outcomes provide learning signals for policy refinement~\cite{fstta,idea}. A central challenge in online navigation adaptation is how to translate individual experiences into lasting improvements in navigation across episodes.


Recent feedback-based methods let navigation agents learn from episode-level success or failure signals: FeedTTA~\cite{feedtta} uses binary feedback for reinforcement learning, and ATENA~\cite{atena} pairs selective feedback queries with outcome-conditioned entropy optimization. Beyond driving the current update, each feedback-labeled episode, with its instruction, observations, actions, and outcome, forms a complete experience that can support later adaptation. This raises a question: \textbf{\textit{as new experiences arrive, how can navigation agents continually reuse the feedback they have already acquired?}}

Human learning offers a perspective on how past experience supports navigation. Complementary learning systems theory suggests that the hippocampus rapidly encodes experiences, whose subsequent reactivation supports gradual knowledge integration in the neocortex~\cite{mcclelland1995there,kumaran2016learning}. As illustrated in Fig.~\ref{fig:head}, a child may mistake a kitchen for a laundry room because both contain a sink and a countertop. Recalling this failure helps avoid repeating the mistake, while remembering a successful visit aids recognition of the correct destination. Repeated recall of both outcomes refines understanding of unfamiliar environments and improves subsequent decisions.

Inspired by this process, we investigate how navigation agents can continually benefit from previously acquired feedback. 
Successful episodes provide evidence for reinforcing effective navigation behavior, while failed episodes provide feedback for discouraging unsuccessful trajectories.
Retaining and revisiting these experiences allows historical feedback to contribute to subsequent policy learning rather than being consumed by a single update. 
However, the policy continually changes during deployment, and a stored experience may already have been reinforced or suppressed by earlier updates.
Repeated replay should therefore account for how the current policy has responded to each experience, while controlling accumulated changes to its pretrained behavior. 
Effective learning from past navigation experiences requires both continual feedback reuse and stable integration of acquired knowledge.

To this end, we propose R$^2$A, an online VLN adaptation framework that continually learns from acquired feedback through episodic replay. 
R$^2$A comprises two complementary modules. 
Success--Failure Memory retains successful and failed navigation episodes with their outcome feedback and retrieves task-relevant experiences to guide subsequent learning.
Prior-Anchored Adaptation adjusts support for recorded trajectories relative to a frozen pretrained policy, reinforcing successes and suppressing failures. Its outcome objective places less weight on experiences whose relative support already aligns with their feedback, while KL regularization constrains behavioral drift. 
Together, these modules enable continual feedback reuse and stable integration of accumulated navigation experiences.

On REVERIE~\cite{reverie} validation unseen, R$^2$A improves DUET's SR/SPL from 46.98\%/33.73\% to \textbf{69.84\%}/\textbf{50.98\%}, with consistent gains on R2R~\cite{vln} and R2R-CE~\cite{r2rce} across five navigation backbones. Ablations confirm the complementary benefits of successful and failed replay and the stabilizing role of prior anchoring. Our contributions are: (1) R$^2$A, an online VLN adaptation framework that continually reuses historical outcome feedback through episodic replay; (2) Success--Failure Memory and reference-relative outcome learning with pretrained-policy regularization; and (3) extensive experiments demonstrating improved navigation performance and stable adaptation across three benchmarks.


%% file: sec/2_related_work_11_11.tex
\begin{figure*}[t]
    \centering
    \includegraphics[width=\linewidth]{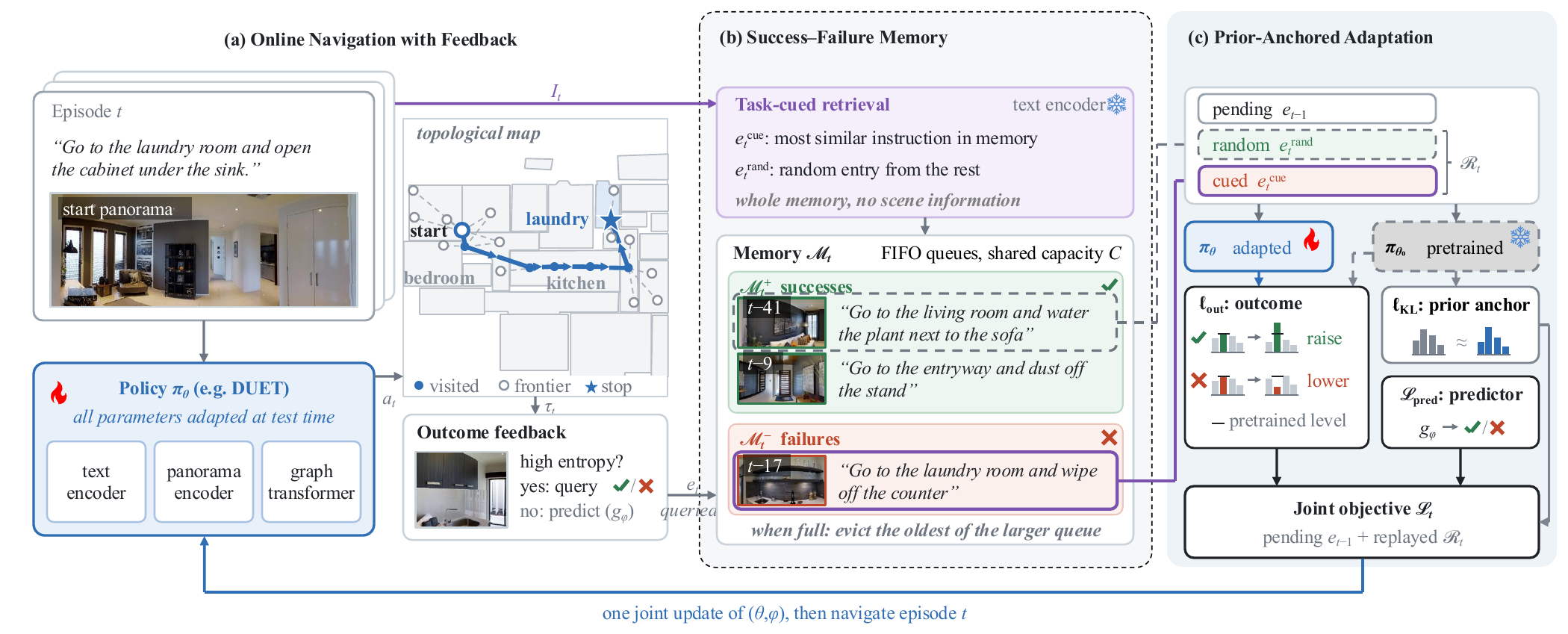}
    \caption{\textbf{Overview of R$^2$A.} (a) The agent navigates using the adapted policy and obtains queried feedback or a self-predicted outcome. (b) \textbf{Success–Failure Memory:} Before navigation, the incoming instruction triggers the retrieval of the most similar experience from the entire memory based on instruction similarity, together with a randomly sampled experience. (c) \textbf{Prior-Anchored Adaptation}: The preceding episode and recalled experiences jointly update the policy through reference-relative outcome learning and pretrained-policy regularization.}
    \label{fig:main}
    \vspace{-2mm}
\end{figure*}

\section{Related Work}
\label{sec:2related}

\textbf{Vision-Language Navigation.}
Vision-and-language navigation (VLN)~\cite{vln} requires agents to make sequential decisions based on natural language instructions and visual observations. Prior work improves instruction--environment alignment and generalization to unseen environments through cross-modal pretraining and large-scale data construction~\cite{tovln,airbert,scalevln,tan2025source}, while history modeling, topological reasoning, and spatial maps support long-horizon exploration and planning~\cite{hamt,duet,gridmm,bevbert,mapgpt}. Iterative VLN~\cite{intervln}, GSA-VLN~\cite{gsavln} and NavGRPO~\cite{navgrpo} further extend this setting to multi-episode navigation in persistent environments, using information from prior exploration to improve subsequent task performance.
These advances highlight the value of past experience. We extend its use from navigation inputs to test-time policy learning by retaining and replaying episodes with outcome feedback.

\noindent\textbf{Test-time Adaptation in VLN.}
Test-time adaptation~(TTA)~\cite{tent,tpt,difftpt} adjusts pretrained models using deployment data to mitigate distribution shifts. Tent~\cite{tent} minimizes prediction entropy for unsupervised adaptation, while subsequent work~\cite{eata,sar,cotta,sharetta} improves reliability through sample selection, stable optimization, and continual adaptation. For vision-language models, entropy-based prompt tuning~\cite{tpt,difftpt} and dynamic caches~\cite{tda,BoostAdapter,mstta,Li_2026_CVPR,soba} offer alternative adaptation mechanisms. These methods primarily target recognition, whereas VLN involves sequential decisions, action-dependent observations, and task outcomes available only at the end of each episode.

Online VLN adaptation methods differ in their use of external outcome feedback. Feedback-free methods, such as FSTTA~\cite{fstta} and IDEA~\cite{idea}, leverage test-time trajectories and historical adaptation knowledge. Feedback-based methods learn from episodic outcomes: FeedTTA~\cite{feedtta} employs reinforcement learning with binary feedback and gradient regularization and further explores llm as feedback oracles, while ATENA~\cite{atena} combines selective feedback querying with outcome-conditioned entropy optimization. R$^2$A retrieves task-relevant historical experiences and learns from their outcomes by adjusting trajectory support relative to a frozen pretrained policy.

\noindent\textbf{Knowledge Retention.} 
Experience replay mitigates forgetting by revisiting historical data, including in continual navigation~\cite{cvln}. Complementarily, knowledge distillation~\cite{kd,fs-kd,wang2025boosting,wang2026bpg} preserves prior knowledge through prediction or feature constraints~\cite{zhang2021self,ding2024lpt}, and transfers policy behavior~\cite{dong2025vita,julg2025refined,zhong2026vla} through action-distribution matching. 
We therefore explore replaying successful and failed navigation episodes to reuse outcome feedback, with a frozen pretrained policy serving as both a reference for outcome learning and an anchor against behavioral drift.

%% file: sec/3_method_11_11.tex
\section{Method}
\label{sec:3method}

We propose R$^2$A to continually reuse deployment feedback through replay of successful and failed experiences. \textbf{S}uccess--\textbf{F}ailure \textbf{M}emory retains feedback-labeled episodes for task-cued replay, while \textbf{P}rior-\textbf{A}nchored \textbf{A}daptation learns from current and past outcomes relative to a frozen pretrained policy and regularizes changes to its action distributions.

\subsection{Problem Setting and Feedback VLN}

\label{sec:3_1_problem}
\noindent\textbf{Problem Setting.}
We consider online VLN in unseen environments. The navigation policy is initialized with pretrained parameters $\theta_0$ and receives navigation instructions sequentially during deployment. In episode $t$, given an instruction $I_t$ and the navigation history $h_{t,k}$, the agent predicts a distribution over the valid action set $\mathcal{A}_{t,k}$, which includes the stop action \texttt{STOP}, and generates a trajectory:
\begin{equation}
    a_{t,k}
    = \arg\max_{a \in \mathcal{A}_{t,k}}
    \pi_{\theta}(a \mid I_t, h_{t,k}),
    \quad
    \tau_t = \{(o_{t,k}, a_{t,k})\}_{k=1}^{L_t},
    \label{eq:navigation_rollout}
\end{equation}
where $h_{t,k}$ and $o_{t,k}$ denote the navigation history and current observation, respectively, and $L_t$ is the trajectory length. The policy is updated after each episode is evaluated, and the updated parameters are carried forward to subsequent episodes. No source-domain data or expert action supervision is required throughout the process.

\noindent\textbf{Feedback VLN.} Following~\cite{feedtta,atena}, we select episodes for feedback queries
based on the mean policy entropy along each trajectory:
\begin{equation}
    q_t = \mathds{1}\left[
    \frac{1}{L_t}\sum_{k=1}^{L_t}
    H\!\left(\pi_{\theta}(\cdot \mid I_t, h_{t,k})\right)
    > \delta
    \right],
    \label{eq:feedback_query}
\end{equation}
where $H(p)=-\sum_a p(a)\log p(a)$ denotes the unnormalized entropy
of the action policy and $\delta$ controls the query frequency.
When $q_t=1$, the feedback query returns an episode-level label
$y_t^{\mathrm{fb}}\in\{+1,-1\}$, indicating success or failure,
respectively.

For episodes without feedback queries, an outcome prediction head
$g_\phi$ estimates the navigation outcome from the mean navigation
state:
\begin{equation}
    \bar{s}_t = \frac{1}{L_t}\sum_{k=1}^{L_t}s_{t,k},
    \qquad
    z_t = g_\phi(\bar{s}_t),
    \label{eq:outcome_prediction}
\end{equation}
where $s_{t,k}$ denotes the state representation produced by the
navigation model at step $k$. The label used for adaptation on
the current episode is:
\begin{equation}
    y_t =
    \begin{cases}
        y_t^{\mathrm{fb}}, & q_t = 1, \\[2pt]
        2\mathds{1}\!\left[\sigma(z_t) > 1/2\right] - 1,
        & q_t = 0,
    \end{cases}
    \label{eq:adaptation_label}
\end{equation}
where $\sigma(\cdot)$ denotes the sigmoid function.

\subsection{Overview}
R$^2$A enables navigation agents to recall and learn from relevant historical experiences before undertaking new tasks. At the beginning of episode $t$, the incoming instruction serves as a cue for retrieving previously acquired feedback from \textbf{S}uccess--\textbf{F}ailure \textbf{M}emory (SFM). The recalled experiences are combined with the outcome of the preceding episode for a single policy update. 
\textbf{P}rior-\textbf{A}nchored \textbf{A}daptation (PAA) adjusts relative support for these recorded trajectories according to their outcomes while regularizing behavioral drift against the pretrained policy.
The adapted policy then executes episode $t$, whose outcome becomes available for the next update. This task-cued update schedule allows historical feedback to guide upcoming navigation without additional optimization steps.
\input{tables/REVERIE}

\subsection{Success--Failure Memory}
\textbf{S}uccess--\textbf{F}ailure \textbf{M}emory (SFM) retains navigation episodes with queried outcome feedback for subsequent adaptation. Successful and failed episodes are stored in separate FIFO queues, preserving both types of experience as navigation outcomes change throughout deployment.

\noindent\textbf{Episode Storage.} Each stored episode is represented as $e=(I_e,\tau_e,y_e)$, where $I_e$ denotes the navigation instruction, $\tau_e$ is the recorded trajectory, and $y_e\in\{+1,-1\}$ is the queried outcome feedback. For efficient retrieval, instruction embeddings are computed using the frozen pretrained text encoder and cached in memory.

\noindent\textbf{Success--Failure Queues.} We maintain separate success and failure queues, $\mathcal{M}_t^+$ and $\mathcal{M}_t^-$, sharing a total capacity of $C$. When the capacity is exceeded, we remove the oldest episode from the larger queue; ties are resolved by removing the earlier-inserted episode across the two queue heads. Only episodes with queried feedback are stored, ensuring that replay uses observed navigation outcomes.




\noindent\textbf{Task-Cued Experience Retrieval.~(TCR)}
Before navigating episode $t$, the agent receives instruction $I_t$ and computes its embedding $z_t$ using the same frozen text encoder. We retrieve the most relevant experience from the entire memory:
\begin{equation}
e_t^{\mathrm{cue}} = \arg\max_{e\in\mathcal{M}_t} z_t^\top z_e,
\end{equation}
where $\mathcal{M}_t=\mathcal{M}_t^+\cup\mathcal{M}_t^-$ contains previously stored successful and failed episodes. This retrieval relies solely on instruction similarity without requiring scene information. To complement task-relevant recall with broader historical experience, we uniformly sample another episode $e_t^{\mathrm{rand}}$ from the remaining memory. The replay set is:
\begin{equation}
\mathcal{R}_t = \{e_t^{\mathrm{cue}},e_t^{\mathrm{rand}}\}.
\end{equation}

When fewer than two memory entries are available, we replay all available entries. Retrieved episodes retain their original feedback, while the current and frozen pretrained policies are evaluated on their recorded histories to compute the adaptation objective.

\subsection{Prior-Anchored Adaptation}
\textbf{P}rior-\textbf{A}nchored \textbf{A}daptation~(PAA) integrates newly acquired outcomes with task-relevant historical feedback using a frozen pretrained policy as a shared reference for outcome learning and behavioral regularization. Given a recorded episode $e$, let $p_{e,k}^{\theta}$ and $p_{e,k}^{\theta_0}$ denote the current and pretrained action distributions at step $k$, evaluated on the same recorded history and valid action candidates.

\noindent\textbf{Reference-Relative Outcome Learning.} As the policy evolves, a stored experience may already have been reinforced or suppressed by previous updates. To account for this change, we measure the policy's relative support for the executed trajectory through a length-normalized log-probability ratio:
\begin{equation}
d_{\theta}(e)
=
\frac{1}{L_e}
\sum_{k=1}^{L_e}
\log
\frac{
p_{e,k}^{\theta}(a_{e,k})
}{
p_{e,k}^{\theta_0}(a_{e,k})
},
\end{equation}
where $a_{e,k}$ is the recorded action and $L_e$ is the episode length. A positive $d_{\theta}(e)$ indicates increased support for the recorded actions relative to the pretrained policy. Given the episode outcome $y_e \in \{+1,-1\}$, we define:
\begin{equation}
\ell_{\mathrm{out}}(e;\theta)
=
-\log \sigma\!\left(y_e d_{\theta}(e)\right),
\end{equation}
where $\sigma$ denotes the sigmoid function. Minimizing this objective increases relative support for successful trajectories and decreases it for failed ones, without explicitly encouraging uniform action distributions after failure.

The gradient takes the form:
\begin{equation}
\nabla_{\theta}\ell_{\mathrm{out}}(e;\theta)
=
-y_e \sigma\!\left(-y_e d_{\theta}(e)\right)
\nabla_{\theta}d_{\theta}(e).
\end{equation}

Thus, the feedback-dependent coefficient decreases as the policy's relative support aligns with the recorded outcome, reducing the weight of experiences whose feedback has already been accommodated. This mechanism introduces no additional temperature or margin parameter. Supervision remains episode-level, without identifying which decisions caused success or failure.

\noindent\textbf{Pretrained Policy Anchoring.} While outcome learning adjusts support for recorded actions, we additionally constrain changes to the full action distributions:
\begin{equation}
\ell_{\mathrm{KL}}(e;\theta)
=
\frac{1}{L_e}
\sum_{k=1}^{L_e}
D_{\mathrm{KL}}\left(
p_{e,k}^{\theta_0}
\parallel
p_{e,k}^{\theta}
\right).
\end{equation}

The frozen policy therefore provides both a reference for measuring feedback-aligned changes and an anchor for regularizing behavioral drift during continual replay.

\subsection{Online Update}
R$^2$A organizes adaptation around the arrival of each new navigation task. After episode $t-1$, its trajectory and outcome are retained for the next update. Upon receiving instruction $I_t$, the agent retrieves the task-cued replay set $\mathcal{R}_t$ from historical memory.

\noindent\textbf{Task-Triggered Adaptation.} The pending episode $e_{t-1}$ and retrieved experiences jointly determine the navigation loss:
\begin{equation}
\mathcal{L}_{t}^{\mathrm{nav}}
=
\ell(e_{t-1};\theta)
+
\overline{\ell}(\mathcal{R}_t;\theta),
\end{equation}
where
\begin{equation}
\ell(e;\theta)
=
\ell_{\mathrm{out}}(e;\theta)
+
\beta\ell_{\mathrm{KL}}(e;\theta),
\end{equation}
and $\overline{\ell}$ denotes the mean over replayed episodes, taking zero when the replay set is empty. Each retrieved episode retains its original queried outcome, while its action distributions and relative support are recomputed under the current policy.

The outcome prediction head is trained on the pending episode through binary cross-entropy. The complete adaptation objective is:
\begin{equation}
\mathcal{L}_t
=
\mathcal{L}_{t}^{\mathrm{nav}}
+
\mathcal{L}_{\mathrm{pred}}(e_{t-1}).
\end{equation}

We perform one joint optimization step before executing episode $t$, updating the policy and outcome prediction head while keeping the pretrained reference frozen. If its feedback was queried, the pending episode is subsequently inserted into the corresponding memory queue. Insertion follows replay selection to prevent the pending episode from also appearing in the replay set for the same update.

\noindent\textbf{Continual Navigation.} The adapted policy executes episode $t$ and obtains either queried feedback or a predicted outcome, which is retained with the trajectory for the next update. The first episode uses the pretrained policy; each subsequent episode is preceded by one joint update. This schedule allows the incoming instruction to guide historical replay without introducing additional optimization steps.

%% file: tables/REVERIE.tex
\begin{table*}[t]
\centering
\caption{\textbf{Comparison with test-time adaptation methods on REVERIE.} \textbf{Bold} indicates the best TTA result within each.}
\label{tab:tta-reverie}
\begingroup
\setlength{\tabcolsep}{3.0pt}
\renewcommand{\arraystretch}{1.0}
\small
\resizebox{\textwidth}{!}{%
\begin{tabular}{lccccccccccccc}
\toprule
\multirow{2}{*}{Method} & \multirow{2}{*}{FB} & \multicolumn{4}{c}{Val Seen} & \multicolumn{4}{c}{Val Unseen} & \multicolumn{4}{c}{Test Unseen} \\
\cmidrule(lr){3-6} \cmidrule(lr){7-10} \cmidrule(lr){11-14}
 & & OSR $\uparrow$ & SR $\uparrow$ & SPL $\uparrow$ & RGSPL $\uparrow$ & OSR $\uparrow$ & SR $\uparrow$ & SPL $\uparrow$ & RGSPL $\uparrow$ & OSR $\uparrow$ & SR $\uparrow$ & SPL $\uparrow$ & RGSPL $\uparrow$ \\
\midrule
\textbf{HAMT}~\cite{hamt} & -- & 47.65 & 43.29 & 40.19 & 25.18 & 36.84 & 32.95 & 30.20 & 17.28 & 33.41 & 30.40 & 26.67 & 13.08 \\
w/ Tent~\cite{tent} & \ding{55} & 46.03 & 43.43 & 40.78 & 25.81 & 32.60 & 30.56 & 28.23 & 14.48 & 25.06 & 23.73 & 21.78 & 10.82 \\
w/ FSTTA~\cite{fstta} & \ding{55} & 48.21 & 42.87 & 39.56 & 24.58 & 36.78 & 32.89 & 30.51 & 17.20 & 33.39 & 30.39 & 26.65 & 13.61 \\
w/ IDEA~\cite{idea} & \ding{55} & 50.67 & 47.33 & 42.13 & 26.82 & 39.87 & 34.92 & 31.52 & 17.76 & 38.14 & 32.81 & 28.52 & 14.45 \\
w/ FeedTTA~\cite{feedtta} & \ding{51} & 62.97 & 55.80 & 49.70 & 31.80 & 40.73 & 35.05 & 31.60 & 17.83 & 38.62 & 34.14 & 29.07 & 14.36 \\
w/ ATENA~\cite{atena} & \ding{51} & 52.92 & 57.34 & 48.08 & 29.60 & 38.85 & 34.00 & 30.96 & 17.51 & 38.19 & 32.55 & 28.38 & 14.32 \\
\rowcolor{black!5}
{w/ $\mathbf{R^2A}$~(Ours)} & \ding{51} & \textbf{65.74} & \textbf{59.41} & \textbf{51.80} & \textbf{34.38} & \textbf{44.17} & \textbf{36.01} & \textbf{32.60} & \textbf{17.94} & \textbf{39.02} & \textbf{35.27} & \textbf{29.96} & \textbf{15.65} \\
\midrule
\textbf{DUET}~\cite{duet} & -- & 73.86 & 71.75 & 63.94 & 51.14 & 51.07 & 46.98 & 33.73 & 23.03 & 56.91 & 52.51 & 36.06 & 22.06 \\
w/ Tent~\cite{tent} & \ding{55} & 73.72 & 71.89 & 64.06 & 50.41 & 51.43 & 47.55 & 33.99 & 23.32 & 57.12 & 52.61 & 36.17 & 22.16 \\
w/ FSTTA~\cite{fstta} & \ding{55} & 75.59 & 75.48 & 65.84 & 52.23 & 56.26 & 54.15 & 36.41 & 23.56 & 58.44 & 53.40 & 36.43 & 22.40 \\
w/ IDEA~\cite{idea} & \ding{55} & 78.45 & 78.24 & 67.74 & 55.07 & 58.51 & 56.92 & 38.03 & 25.47 & 58.91 & 55.12 & 39.84 & 24.52 \\
w/ FeedTTA~\cite{feedtta} & \ding{51} & 86.16 & 84.19 & 75.54 & 60.32 & 71.60 & 66.49 & 45.38 & 30.75 & 58.76 & 53.58 & 37.66 & 24.10 \\
w/ ATENA~\cite{atena} & \ding{51} & 85.52 & 84.33 & 74.31 & 59.99 & \textbf{71.88} & 68.11 & 45.82 & 31.26 & 57.74 & 54.28 & 40.70 & 25.01 \\
\rowcolor{black!5}
{w/ $\mathbf{R^2A}$~(Ours)} & \ding{51} & \textbf{87.93} & \textbf{85.87} & \textbf{78.19} & \textbf{62.13} & 71.46 & \textbf{69.84} & \textbf{50.98} & \textbf{33.27} & \textbf{60.45} & \textbf{58.44} & \textbf{42.47} & \textbf{26.81} \\
\midrule
\textbf{GOAT}~\cite{goat} & -- & 82.36 & 80.74 & 73.44 & 58.82 & 57.97 & 53.82 & 37.52 & 27.00 & 61.44 & 57.72 & 40.53 & 26.70 \\
w/ Tent~\cite{tent} & \ding{55} & 82.43 & 80.74 & 73.47 & 58.75 & 57.68 & 53.51 & 37.49 & 26.99 & 62.00 & 57.28 & 39.82 & 26.97 \\
w/ FSTTA~\cite{fstta} & \ding{55} & 82.36 & 80.74 & 73.42 & 58.82 & 57.94 & 53.79 & 37.50 & 26.95 & 62.35 & 57.52 & 39.49 & 26.82 \\
w/ ATENA~\cite{atena} & \ding{51} & 85.03 & 83.35 & 76.45 & 61.60 & 70.29 & 67.66 & 53.15 & 39.80 & 64.26 & 62.03 & 46.82 & 31.54 \\
\rowcolor{black!5}
{w/ $\mathbf{R^2A}$~(Ours)} & \ding{51} & \textbf{88.69} & \textbf{87.63} & \textbf{79.90} & \textbf{63.86} & \textbf{72.62} & \textbf{69.58} & \textbf{56.09} & \textbf{40.69} & \textbf{67.07} & \textbf{64.79} & \textbf{48.83} & \textbf{34.27} \\
\bottomrule
\end{tabular}%
}
\endgroup
\end{table*}

%% file: sec/4_experiment_11_11.tex

\input{tables/R2R-CE}

\section{Experiment}
\label{sec:4eperiment}

\subsection{Experimental Setup}
\label{sec:4_1setup}
\textbf{Datasets \& Metrics.}
We evaluate R$^2$A on three VLN benchmarks: REVERIE~\cite{reverie}, R2R~\cite{vln}, and R2R-CE~\cite{r2rce}. REVERIE requires navigation and target object identification from high-level instructions, while R2R involves following fine-grained route instructions. R2R-CE extends R2R to continuous environments with low-level actions. Following previous evaluation protocols   ~\cite{vln,fstta}, navigation succeeds when the agent stops at a viewpoint where the target object is visible on REVERIE, or within 3 meters of the goal on R2R and R2R-CE. We report OSR, SR, SPL, and RGSPL for REVERIE, and TL, NE, SR, and SPL for R2R and R2R-CE. Evaluation covers validation seen and unseen splits, along with the REVERIE test unseen split.

\noindent\textbf{Baselines.}
We evaluate R$^2$A on five pretrained VLN backbones: HAMT~\cite{hamt}, DUET~\cite{duet}, BEVBert~\cite{bevbert}, GOAT~\cite{goat}, and ETPNav~\cite{etpnav}. We use HAMT, DUET, and GOAT on REVERIE; DUET, BEVBert, and GOAT on R2R; and ETPNav and BEVBert on R2R-CE. We compare against feedback-free TTA methods (Tent~\cite{tent}, FSTTA~\cite{fstta}, IDEA~\cite{idea}) and feedback-based methods (FeedTTA~\cite{feedtta}, ATENA~\cite{atena}). Baseline results are taken from the original publications.

\input{tables/R2R}

\noindent\textbf{Implementation Details.}
We follow the online adaptation protocol, updating the policy after each episode while retaining parameters across the stream. Following ATENA~\cite{atena}, feedback is queried at an entropy threshold of $\delta=0.1$, with unqueried outcomes self-predicted. Success–Failure Memory stores only feedback-labeled episodes with capacity $C=300$ and replays up to $R=2$ episodes per update. We use an anchoring coefficient of $\beta=0.1$ with a frozen pretrained reference policy. Results are averaged over 3 episode orders for all datasets. Further implementation details are provided in the Sec.~6.

\begin{figure*}[t]
    \centering
    \includegraphics[width=\linewidth]{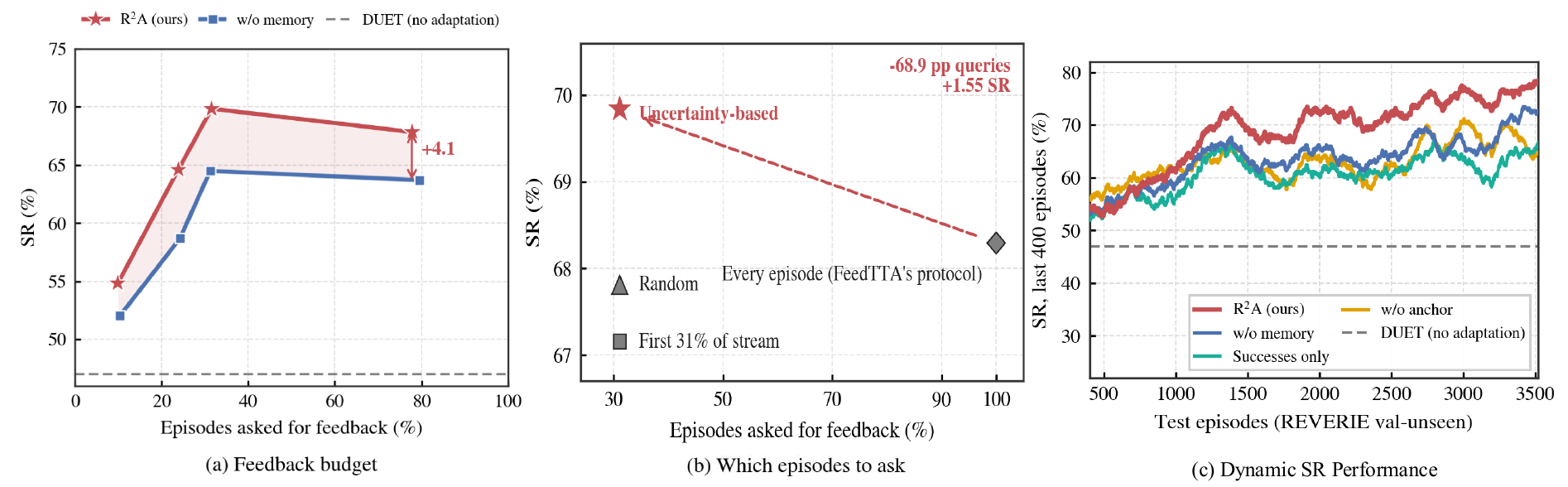}
    \caption{\textbf{Feedback efficiency and online adaptation on REVERIE val-unseen.} (a) Performance under different feedback budgets. (b) Comparison of feedback query strategies. (c) Evolution of navigation SR throughout online adaptation.}
    \label{fig:dynamic}
    \vspace{-2mm}
\end{figure*}

\subsection{Main Navigation Results}
\label{sec:4_2main_results}

\noindent\textbf{REVERIE.} As shown in Table~\ref{tab:tta-reverie}, R$^2$A consistently improves navigation performance across HAMT, DUET, and GOAT on both seen and unseen environments. On the validation unseen split, R$^2$A improves DUET's SR from 46.98\% to \textbf{69.84\%} and SPL from 33.73\% to \textbf{50.98\%}, outperforming ATENA's 68.11\% SR and 45.82\% SPL. Notably, RGSPL also increases from 23.03\% to \textbf{33.27\%}, indicating improved performance in navigation tasks requiring both successful navigation and target object grounding. Similar improvements are observed with GOAT, where R$^2$A achieves \textbf{69.58\%} SR and \textbf{56.09\%} SPL, surpassing ATENA by 1.92\% and 2.94\%, respectively. On the test unseen split, R$^2$A further improves DUET's SR and SPL to \textbf{58.44\%} and \textbf{42.47\%}, compared with 54.28\% and 40.70\% achieved by ATENA. These consistent gains across 3 backbones and evaluation splits demonstrate the effectiveness of continually \textbf{incorporating historical feedback into online policy learning}, enabling navigation agents to progressively improve their performance in unseen environments.

\noindent\textbf{R2R \& R2R-CE.} Tables~\ref{tab:tta-r2r} and~\ref{tab:tta-r2r-ce} evaluate R$^2$A in discrete and continuous navigation. On R2R validation unseen, R$^2$A improves DUET's SR/SPL from 72\%/60\% to \textbf{77\%/69\%}, surpassing ATENA's 75\%/66\%. BEVBert similarly improves from 75\%/64\% to \textbf{77\%/69\%,} while the stronger GOAT backbone reaches \textbf{80.42\%/71.30\%} from 77.91\%/67.34\%, outperforming existing TTA methods. On R2R-CE validation unseen, R$^2$A raises ETPNav's SR/SPL from 57\%/49\% to \textbf{60\%/51\%}, while BEVBert achieves \textbf{63\%/53\%}, exceeding IDEA's 62\%/52\%. These consistent gains across backbones and navigation settings demonstrate the applicability of \textbf{historical feedback reuse to both discrete planning and continuous navigation}.

\subsection{Ablation Studies and Analysis}
\label{sec:4_3ablation}

\begin{figure}[t]
    \centering
    \includegraphics[width=\linewidth]{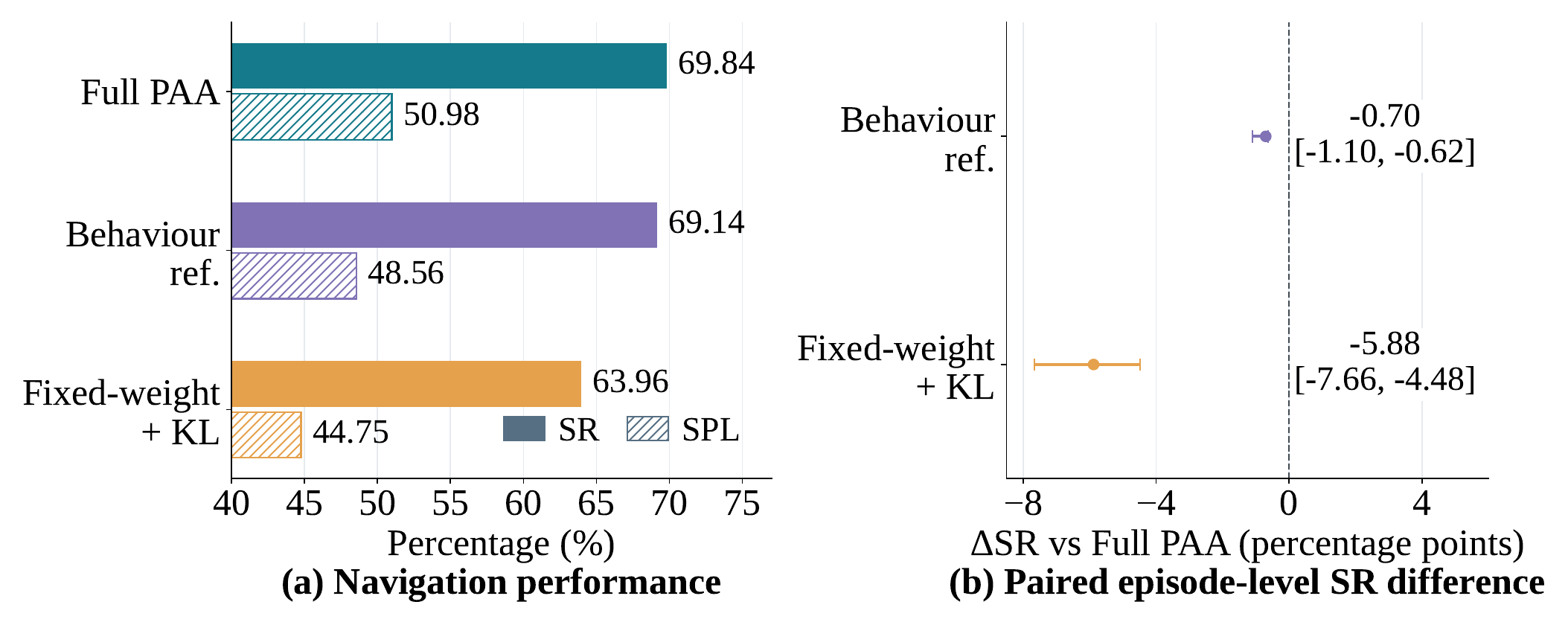}
\caption{\textbf{Analysis of PAA components on REVERIE val-unseen.} (a) Navigation performance under different outcome weighting and reference choices. (b) Paired SR differences relative to full PAA.}
    \label{fig:paa-ablation}
    \vspace{-2mm}
\end{figure}

\begin{figure*}[t]
    \centering
    \includegraphics[width=\linewidth]{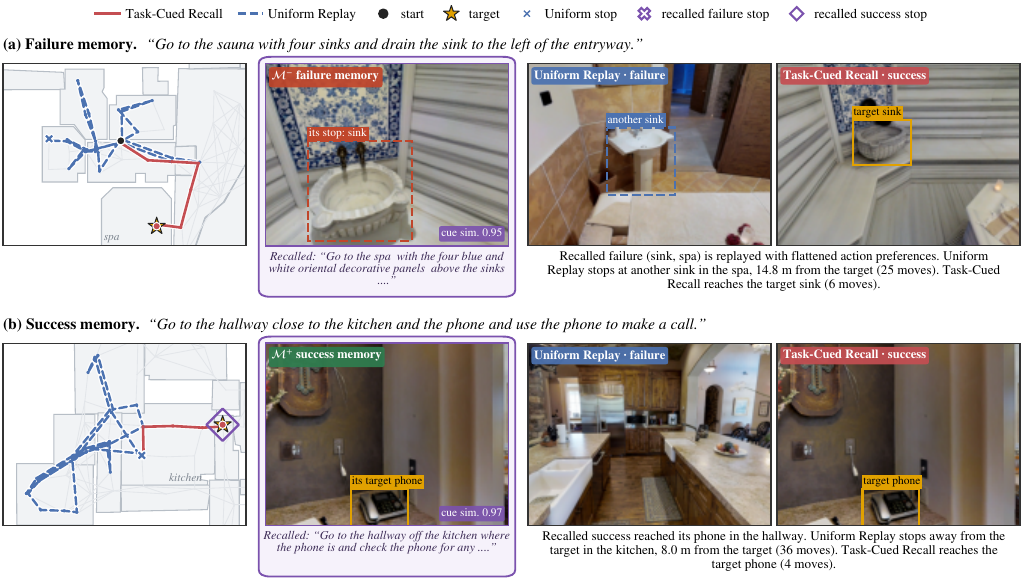}
\caption{\textbf{Qualitative examples of task-cued recall on REVERIE val-unseen.} (a) Failure memory: recalling a previously failed navigation experience helps the agent avoid stopping at an incorrect sink. (b) Success memory: recalling a successful experience guides the agent toward the target phone. In both cases, Task-Cued Recall reaches the intended destination, while uniform replay fails.}
    \label{fig:qualitative}
    \vspace{-2mm}
\end{figure*}

\noindent\textbf{Effect of Historical Feedback Replay.} Table~\ref{tab:component-ablation} examines the contribution of historical feedback to online adaptation. Removing memory reduces SR from 69.84\% to 64.49\% on REVERIE and from 77\% to 74\% on R2R. On REVERIE, success-only and failure-only replay achieve 66.19\% and 53.78\% SR, respectively, while combining both improves SR to \textbf{69.84\%}. Excluding failure feedback from both current updates and memory further reduces SR to 59.43\%. These results demonstrate the complementary\textbf{ roles of successful and failed experiences} in continual adaptation.


\input{tables/components}


\noindent\textbf{Effect of Prior-Anchored Adaptation.}
PAA enables effective outcome-driven learning while limiting behavioral drift during online adaptation. Removing the pretrained policy anchor reduces REVERIE SR from 69.84\% to 56.74\% and increases the standard deviation across episode orders from 0.70 to 7.34, indicating both \textbf{degraded performance and greater sensitivity to episode ordering}. The increased fluctuations in Fig.~\ref{fig:dynamic}(c), together with the SR drop from 77\% to 73\% on R2R, further support the anchor's role in stabilizing adaptation. Beyond the presence of an anchor, \textbf{the adaptation objective also matters}: full PAA outperforms the behavior-reference variant (69.84\% vs. 69.14\% SR; 50.98\% vs. 48.56\% SPL) and fixed-weight KL regularization (63.96\% SR, 44.75\% SPL), as shown in Fig.~\ref{fig:paa-ablation}. Paired episode-level comparisons corroborate these gains, suggesting that PAA's benefits extend beyond simply imposing a policy constraint. Together, these results show that PAA supports \textbf{learning from outcome feedback while maintaining stable navigation behavior}.


\noindent\textbf{Effect of Feedback Querying.} Fig.~\ref{fig:dynamic}(b) compares different feedback acquisition strategies. Uncertainty-based querying achieves 69.84\% SR using feedback from only 31.1\% of episodes, outperforming random and sequential querying under the same budget. Compared with querying every episode, it achieves 1.55\% higher SR with 68.9\% fewer feedback queries.

\input{tables/recall_analysis}
\noindent\textbf{Effect of Task-Cued Recall.}
Table~\ref{tab:recall-strategies} compares experience replay strategies. TCR consistently outperforms uniform replay, improving SR by \textbf{3.30\%} on REVERIE and \textbf{3\%} on R2R. Globally shuffling instruction cues reduces REVERIE SR to 66.82\%, close to uniform replay, highlighting the importance of \textbf{instruction--experience correspondence}. TCR also surpasses scene-based recall, demonstrating the benefit of instruction-guided retrieval without scene information. Excluding same-goal experiences causes only a modest performance drop, suggesting that TCR benefits from \textbf{task-relevant feedback beyond repeated goals}.

\noindent\textbf{Qualitative Analysis.} Fig.~\ref{fig:qualitative} illustrates how task-cued recall improves navigation by reusing relevant historical feedback. In Fig.~\ref{fig:qualitative}(a), recalling a failed experience helps the agent avoid stopping at an incorrect sink, reaching the target in 6 moves instead of the 25 moves taken by the unsuccessful uniform-replay agent. In Fig.~\ref{fig:qualitative}(b), recalling a successful experience guides the agent to the target phone in 4 moves, while uniform replay fails after 36 moves. These examples demonstrate the complementary value of recalling past successes and failures for subsequent navigation.


\noindent\textbf{Supplementary Experiments \& Analysis.} The supplement details implementation (Sec.~6), hyperparameter sensitivity (Sec.~7.1), feedback effects (Secs.~7.2, 7.3, 7.5), memory and anchoring designs (Secs.7.4, 7.6, 7.7), efficiency (Sec.~7.8), feedback robustness (Secs.~8.1, 8.2), and integration with other adaptation methods (Secs.~8.3, 8.4).

%% file: tables/R2R-CE.tex
\begin{table}[t]
\centering
\caption{\textbf{Comparison with VLN-TTA methods on R2R-CE.} 
}
\label{tab:tta-r2r-ce}
\begingroup
\setlength{\tabcolsep}{1.5pt}
\renewcommand{\arraystretch}{1.00}
\small
\begin{tabular}{lcccccccc}
\toprule
\multirow{2}{*}{Method}
& \multicolumn{4}{c}{Val Seen}
& \multicolumn{4}{c}{Val Unseen} \\
\cmidrule(lr){2-5} \cmidrule(lr){6-9}
& TL$\downarrow$ & NE$\downarrow$ & SR$\uparrow$ & SPL$\uparrow$
& TL$\downarrow$ & NE$\downarrow$ & SR$\uparrow$ & SPL$\uparrow$ \\
\midrule
\textbf{ETPNav}~\cite{etpnav} & 11.78 & 3.95 & 66 & 59 & 11.99 & 4.71 & 57 & 49 \\
w/ Tent~\cite{tent} & 11.36 & 3.94 & 66 & 59 & 11.57 & 4.74 & 57 & 49 \\
w/ FSTTA~\cite{fstta} & 11.35 & 3.93 & 66 & 59 & 11.57 & 4.77 & 57 & 49 \\
w/ IDEA~\cite{idea} & 10.87 & \textbf{3.82} & 68 & \textbf{62} & \textbf{11.17} & \textbf{4.47} & 59 & \textbf{51} \\
w/ FeedTTA~\cite{feedtta} & 10.88 & 3.85 & 67 & 61 & 11.99 & \textbf{4.47} & 58 & 50 \\
w/ ATENA~\cite{atena} & 10.81 & 3.86 & 67 & 61 & 12.89 & 4.53 & 58 & 49 \\
\rowcolor{black!5}
{w/ $\mathbf{R^2A}$~(Ours)} & \textbf{10.69} & 3.88 & \textbf{69} & \textbf{62} & 11.31 & 4.49 & \textbf{60} & \textbf{51} \\
\midrule
\textbf{BEVBert}~\cite{bevbert} & 13.98 & 3.77 & 68 & 60 & 13.27 & 4.57 & 59 & 50 \\
w/ Tent~\cite{tent} & 12.74 & 3.35 & 70 & 62 & 13.29 & 4.61 & 59 & 49 \\
w/ FSTTA~\cite{fstta} & 14.07 & 4.11 & 69 & 60 & 13.11 & 4.39 & 60 & 51 \\
w/ IDEA~\cite{idea} & 12.27 & \textbf{3.04} & \textbf{73} & \textbf{64} & 12.67 & {4.26} & {62} & {52} \\
w/ FeedTTA~\cite{feedtta} & 13.54 & 3.08 & \textbf{73} & 63 & 16.15 & 4.33 & 61 & 50 \\
w/ ATENA~\cite{atena}& 11.31 & 3.24 & 71 & \textbf{64} & 13.48 & 4.50 & 60 & 51 \\
\rowcolor{black!5}
{w/ $\mathbf{R^2A}$~(Ours)} & \textbf{11.06} & 3.10 & \textbf{73} & \textbf{64} & \textbf{12.49} & \textbf{4.21} & \textbf{63} & \textbf{53} \\
\bottomrule
\end{tabular}
\endgroup
\end{table}

%% file: tables/R2R.tex
\begin{table}[t]
\centering
\caption{\textbf{Comparison with VLN-TTA methods on R2R.} 
}
\label{tab:tta-r2r}
\begingroup
\setlength{\tabcolsep}{1.1pt}
\renewcommand{\arraystretch}{1.0}
\small
{%
\begin{tabular}{lcccccccc}
\toprule
\multirow{2}{*}{Method}
& \multicolumn{4}{c}{Val Seen}
& \multicolumn{4}{c}{Val Unseen} \\
\cmidrule(lr){2-5} \cmidrule(lr){6-9}
& TL$\downarrow$ & NE$\downarrow$ & SR$\uparrow$ & SPL$\uparrow$
& TL$\downarrow$ & NE$\downarrow$ & SR$\uparrow$ & SPL$\uparrow$ \\
\midrule
\textbf{DUET}~\cite{duet} & 12.33 & 2.28 & 79 & 73 & 13.94 & 3.31 & 72 & 60 \\
w/ Tent~\cite{tent} & 12.17 & 2.38 & 78 & 72 & 13.78 & 3.42 & 72 & 60 \\
w/ FSTTA~\cite{fstta} & 13.39 & 2.25 & 79 & 73 & 14.64 & 3.03 & 75 & 62 \\
w/ IDEA~\cite{idea} & \textbf{11.23} & \textbf{2.03} & 81 & 76 & 12.47 & 2.91 & 76 & 67 \\
w/ FeedTTA~\cite{feedtta} & 11.49 & 2.09 & 80 & 75 & 13.52 & 2.95 & 75 & 65 \\
w/ ATENA~\cite{atena} & 11.27 & 2.18 & 80 & 75 & \textbf{12.31} & 2.90 & 75 & 66 \\
\rowcolor{black!5}
{w/} $\mathbf{R^2A}$~(Ours) & 11.25 & 2.05 & \textbf{83} & \textbf{78} & 13.80 & \textbf{2.83} & \textbf{77} & \textbf{69} \\
\midrule
\textbf{BEVBert}~\cite{bevbert} & 13.56 & 2.17 & 81 & 74 & 14.55 & 2.81 & 75 & 64 \\
w/ Tent~\cite{tent} & 12.68 & 2.36 & 80 & 74 & 13.14 & 2.93 & 74 & 63 \\
w/ FSTTA~\cite{fstta} & 12.28 & 2.31 & 80 & 75 & 13.96 & 2.89 & 74 & 63 \\
w/ IDEA~\cite{idea} & 10.89 & 2.15 & 83 & \textbf{79} & 12.03 & 2.53 & 76 & 68 \\
w/ FeedTTA~\cite{feedtta} & 11.88 & 2.17 & 82 & 77 & 12.24 & 2.77 & 75 & 66 \\
w/ ATENA~\cite{atena} & 10.79 & 2.26 & 82 & 78 & 12.22 & 2.78 & 76 & 68 \\
\rowcolor{black!5}
{w/} $\mathbf{R^2A}$~(Ours) & \textbf{10.63} & \textbf{2.13} & \textbf{84} & \textbf{79} & \textbf{12.01} & \textbf{2.41} & \textbf{77} & \textbf{69} \\
\midrule
\textbf{GOAT}~\cite{goat} & 11.87 & 1.70 & 84.52 & 79.60 & 13.43 & 2.33 & 77.91 & 67.34 \\
w/ FSTTA~\cite{fstta} & 11.67 & 1.65 & 84.92 & 80.08 & 13.26 & 2.32 & 77.99 & 67.48 \\
w/ ATENA~\cite{atena} & 11.66 & 1.64 & 85.01 & 80.13 & 12.52 & 2.27 & 79.01 & 69.30 \\
\rowcolor{black!5}
{w/} $\mathbf{R^2A}$~(Ours) & \textbf{11.58} & \textbf{1.56} & \textbf{85.27} & \textbf{80.19} & \textbf{11.75} & \textbf{2.14} & \textbf{80.42} & \textbf{71.30} \\
\bottomrule
\end{tabular}%
}
\endgroup
\end{table}

%% file: tables/components.tex

\begin{table}[t]
\centering
\caption{\textbf{Effects of feedback replay and prior anchoring.}
Results use DUET on REVERIE and R2R val-unseen.
Success-only memory retains failure feedback in current updates; w/o failure feedback excludes it from both current updates and replay.}
\label{tab:component-ablation}
\begingroup
\setlength{\tabcolsep}{3.6pt}
\renewcommand{\arraystretch}{1.01}
\small
\begin{tabular}{lcccc}
\toprule
\multirow{2}{*}{Variant}
& \multicolumn{2}{c}{REVERIE}
& \multicolumn{2}{c}{R2R} \\
\cmidrule(lr){2-3} \cmidrule(lr){4-5}
& SR$\uparrow$ & SPL$\uparrow$
& SR$\uparrow$ & SPL$\uparrow$ \\
\midrule
DUET (no adaptation)
& 46.98 & 33.73 & 72 & 60 \\
\midrule
\rowcolor{black!5}
$\mathbf{R^2A}$ (full)
& $\mathbf{69.84 \pm 0.70}$
& \textbf{50.98}
& \textbf{77}
& \textbf{69} \\
\midrule
\quad w/o memory
& $64.49 \pm 0.86$ & 48.55 & 74 & 66 \\
\quad w/o anchor
& $56.74 \pm 7.34$ & 39.45 & 73 & 63 \\
\quad Success-only memory
& $66.19 \pm 0.27$ & 46.51 & 74 & 65 \\
\quad Failure-only memory
& $53.78 \pm 1.18$ & 42.55 & 73 & 64 \\
\quad w/o failure feedback
& $59.43 \pm 0.72$ & 42.57 & 73 & 62 \\
\bottomrule
\end{tabular}
\endgroup
\vspace{-3mm}
\end{table}

%% file: tables/recall_analysis.tex
\begin{table}[t]
\centering
\caption{\textbf{Comparison of replay strategies on val unseen.}}
\label{tab:recall-strategies}
\begingroup
\setlength{\tabcolsep}{8pt}
\renewcommand{\arraystretch}{1.05}
\small
\begin{tabular}{lcccc}
\toprule
\multirow{2}{*}{Method} & \multicolumn{2}{c}{REVERIE} & \multicolumn{2}{c}{R2R} \\
\cmidrule(lr){2-3} \cmidrule(lr){4-5}
 & SR $\uparrow$ & SPL $\uparrow$ & SR $\uparrow$ & SPL $\uparrow$ \\
\midrule
Uniform Replay & 66.54 & 48.12 & 74 & 65 \\
Global Shuffled-Cue & 66.82 & 48.34 & 74 & 65 \\
Scene-Based Recall & 67.87 & 49.91 & 75 & 67 \\
TCR w/o Same-Goal & 69.26 & 50.40 & 76 & 68 \\
\rowcolor{black!5}\textbf{TCR (Ours)} & \textbf{69.84} & \textbf{50.98} & \textbf{77} & \textbf{69} \\
\bottomrule
\end{tabular}
\endgroup
\vspace{-2mm}
\end{table}

%% file: sec/5_conclusion_11_11.tex
\section{Conclusion}

We proposed R$^2$A, an online VLN adaptation framework that continually reuses acquired feedback through task-cued episodic replay. Success–Failure Memory retains successful and failed experiences and retrieves task-relevant feedback to guide subsequent navigation, while Prior-Anchored Adaptation learns from these outcomes by adjusting trajectory support relative to a frozen pretrained policy while regularizing behavioral drift. Experiments across 3 benchmarks and 5 navigation backbones demonstrate consistent performance improvements, while further analyses validate the benefits of task-relevant replay and stable adaptation.

\section*{LLM Usage Statement}
OpenAI ChatGPT was used to assist with language editing, grammatical refinement, and improving the clarity of the manuscript. All scientific content, experimental results, and final revisions were reviewed and verified by the authors.